\documentclass[10pt,a4paper]{article}
\usepackage[latin1]{inputenc}
\usepackage{amsmath}
\usepackage{amsfonts}
\usepackage{amssymb}
\usepackage{hyperref}
\usepackage{graphics}
\begin{document}
\author{Lin Duan\\
Department of Computer Science\\
Yunnan University\\
lin.duan.zhang@hotmail.com}
\title{Exploration of object recognition from 3D point cloud}
\maketitle

\section{Introduction of the project}
The project I am currently working on is about object detection and recognition from street view LiDAR point cloud. The challenge of this project is very evident that the computational cost and speed are the most bottle neck. Almost all existing method conduct a segmentation based object recognition schema\cite{zhang20083d, zhang2015learning, zhang2014alignment}. However, segmentation of point cloud itself is not a solved problem and it is not a practical thing to do if the processing speed is a critical consideration. 

Our proposed method conduct object recognition and detection does not require segmentation of objects to be done in advance. In fact, we prefer to consider this problem as a template matching problem that we will generate point cloud templates for objects that we set as detection and recognition targets\cite{zhang2013distinguishable2, zhang2013structure2, zhang2014learning2}. So that during detection and recognition, we compare templates with all candidates. Recognition or say classification then is done based on results of the comparison.

Recognition of the objects can be much easier done if segmentation information is provided\cite{zhang2014learning, zhang2015learning1}, which actually in turn offers semantic information of each part and relations between the parts. Since our method does not require segmentation of the objects prior to recognition\cite{zhang2012learning, zang2015learning}.  To provide semantics of each parts, we first gridded objects to cells\cite{zhang2016cgmos}, then we compute object feature as a combination of spatial shape information and a statistical model we enforced on gridded object. 

\section{Current progress}
In this semester, I make a lot of progress on current project. I will use this section to introduce these progress respectively. 

\subsection{Candidate location selection}
Our template based detection schema require a scan of input data using object template. So, it would be helpful if we can rule out locations that less likely have target objects. The goal in detection is to minimize the missing rate of the target objects while get rid of obvious outliers as much as possible. The key idea is to use several weak classifiers in every step and refine the classification step by step.

I have done some simple filtering techniques to detect ROI. These techniques based on project/push down point clouds to regular divided ground tiles. The results looks promising for these filtering techniques, at least all interesting objects are included. But to further refine ROI region, I propose to search valid ROI in feature space of the ground tile structure. So, for each specific type of object, I will compute a cluster center or fit a distribution in feature space. For candidate ROI tile, I compute the corresponding features for the tile, and put this feature in feature space I just modelled. We either can threshold the distance between cluster center and tile's feature or only make K nearest tiles in feature space as valid in the next step. 

\begin{figure}
\centerline{
\begin{tabular}{c} 
  \resizebox{0.8\textwidth}{!}{\rotatebox{0}{
  \includegraphics{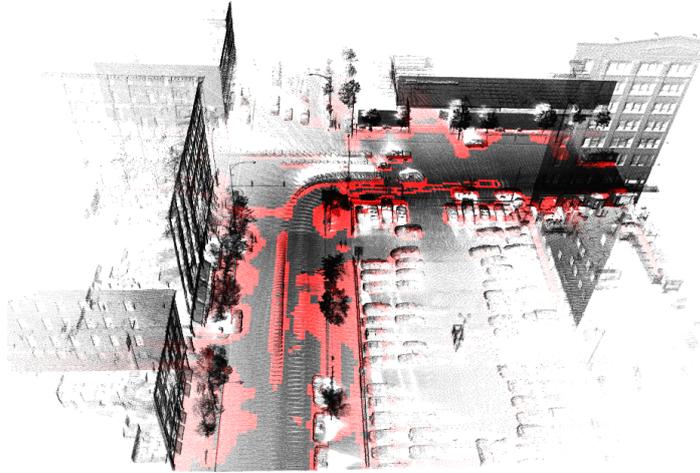}}}
  \\
\end{tabular}}
\caption{Current result of ROI detection}
\label{fig: modified distance.}
\end{figure} 

\subsection{Preprocessing of alignment}
When aligning the objects, the correspondent parts among object should overlap as much as possible. So, it is necessary to get geometric information about each part of the object. Most of previous approaches conduct a segmentation to separate the parts of the object. However, it is costy to have a meaningful segmentation by running algorithm such as graph cut, hierarchical processing and so on. In stead, I consider this process as a point clustering problem where points with similar geometric context information are grouped together.

Since the speed is one of concern in our method, the computation of the features should be as efficient as possible while still contain rich information of the context. 

In this step, I compute spin images for each point. However, straight comparison based on spin image fail to catch enough correlation among similar images. To better explore and compute the correlation among spin images. I tried three different methods; all of three methods are based on one idea that I want to encode each spin image as a combination of a set of codes. 

The first method I try is a method based on sparse coding which can learn detailed features of the image, however, the method I try requires to solve a quadratic problem which is slow.

In the second method, similar to the first method, it is also require a generation of code book, however, this time, the code is given by principle component analysis. Since I want to catch the details of the spin image. Given a spin image with $31\times 16$ resolution, for each pixel, I take a $11\times 11$ patch from it. A code book than is computed by conducting a PCA on all patches from training data. The top 30 eigenvectors then are used in the code book. This method works, but since each spin image is encoded based on very small size of codes, the method is sensitive to outliers and noise.

In the third method, instead of using small patch to train code book, I train a code book using spin image itself. Given a spin image of $31\times 16$, I take top 30 eigenvector as codes. Then for each spin image, it can be encoded as the coefficient of projection on each eigenvector. This method generate best results among above three approaches.

\begin{figure}
\centerline{
\begin{tabular}{cc} 
  \resizebox{0.5\textwidth}{!}{\rotatebox{0}{
  \includegraphics{Pic/PDF/car.pdf}}}
  &
  \resizebox{0.5\textwidth}{!}{\rotatebox{0}{
  \includegraphics{Pic/PDF/lowobj.pdf}}}
  \\
  car & garbage can
  \\
  \resizebox{0.5\textwidth}{!}{\rotatebox{0}{
  \includegraphics{Pic/PDF/pedestrain.pdf}}}
  &
  \resizebox{0.5\textwidth}{!}{\rotatebox{0}{
  \includegraphics{Pic/PDF/strlight.pdf}}}
  \\
  pedestrain & street light
\end{tabular}}
\caption{Car}
\label{fig: toy sample}
\end{figure} 

\subsection{Alignment of objects of same type}
Last week we discussed about aligning object by segmented parts, and come up with an idea using message passing like mechanism to align objects.

I search for relative literature. There is one paper\cite{johnson1997surface} talked about aligning object using parts and the optimal global alignment is decided by alignment of part giving least error. I also look at several papers talking about shape correspondence analysis from a group of objects\cite{wang2012active, van2011survey, van2013co, golovinskiy2009consistent, kim2012exploring, shao2013interpreting, kim2013learning, van2011prior, sidi2011unsupervised}. Most of these papers have some thing in common when they analysis correspondence. First, the correspondence is done mostly in feature space using clustering algorithm which is similar as ours. Second, several papers use ICP to align object before they can build correspondence. For ICP in these papers, they use the normal ICP.

Speaking of our problem, we have already segment objects to parts, to do message passing part based ICP, we have to build contact graph first, where touch parts are connected by an edge in this graph. Since we are dealing alignment problem for real world objects which are up-ward oriented already. So, the transformation has 4 degree of freedom only that we only allow: horizontal translation, z direction translation, and 2D rotation around z axis. Then for segmented parts of each object, I proposed to restrict their transformation to one or several of 4 degree of freedom movements. This idea comes from observation that some part of object is more sensitive to some movement and can produce large error in some certain movement if misaligned. And the good thing to have this restriction in contact graph is in that it provide us a way to deal with message passing between graph nodes.

Another thing is to align a group of objects, we need to figure out that how we choose alignment target from a group of objects. One option is to manually choose a almost perfect model from the group, however, the final alignment is bias towards the target model. In \cite{kim2012exploring}, authors propose to build a similarity matrix among all objects, and from that matrix, they derive a minimum spanning tree(MST). The alignment only happens between nodes connected by the tree edge. I think we can borrow the same idea in our work. But this method is very expensive that we have to compute similarity matrix every time when all nodes in current MST is accessed.

I also propose another method when align a pair of objects after all settings mentioned above are done. At this point, we need to align object by their parts, however, we have no idea which part should we start first and which one is the next. So, we have to have some pre-knowledge about the correspondence quality between correspondent parts of two objects. Such knowledge is given by compare the feature correlation between the parts. 

To have a baseline, we implement merging using ICP first. In a group of objects, we want to start merging objects which are most similar to each other. So, to get the similarity measurement, I compute five features each object. These five features characterize different shape of the object, such as the thickness of the object, the distribution of points respect to some direction and so on. Then, the similarity will be computed between each pair of objects in the group which yields a similarity matrix for the objects in the group. I use Earth Mover's Distance to compute the distance between two histogram features, and the final similarity is given by averaging measurement on five features. 

Given similarity matrix, I represent each object using a set I employee a divide and conquer strategy in our alignment algorithm. Each iteration, two sets with shortest distance between them will be merged. When merging two sets of objects, the transformation matrix is computed based on alignment of two most similar objects in the two groups. 

We get some preliminary results by using above schema, we show the results in the following figure:

\begin{figure}
\centerline{
\begin{tabular}{c} 
  \resizebox{0.8\textwidth}{!}{\rotatebox{0}{
  \includegraphics{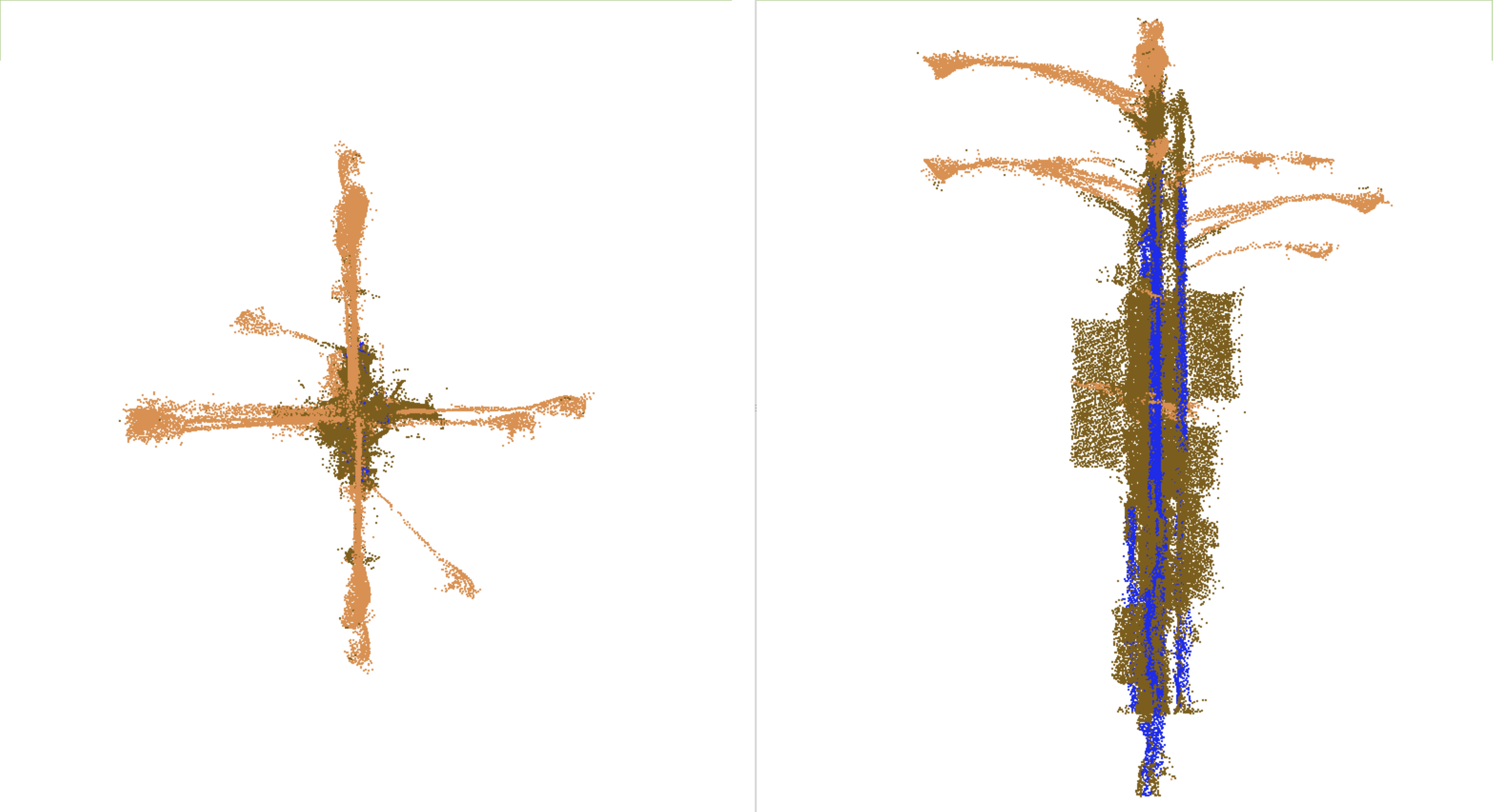}}}
  \\
\end{tabular}}
\caption{Current result of ROI detection}
\label{fig: modified distance.}
\end{figure} 

\subsection{Hierarchical Shape Distribution}
Given a point cloud $P=\{p_i\}_{i=1}^n$, the original shape distribution feature compute several geometry measurements for all point combinations of 2-points or 3-points and use histograms to save the measurements.
\\
\\
\noindent \textit{2-points features:} D2 (Distance between any two points).\\
\noindent \textit{3-points features:} A3 (Area of triangle composed by any three points), T3 (Volume of tetrahedron composed by any four points). R3 (Radius of inscribed circle of triangle composed by any three points).
\\
\\
When the number of points in a point cloud is over some value, computer crashes due to the memory issue, and also the computational cost is very high.

So, we build a hierarchical structure to limit the computation of shape distribution feature in each level and component of the structure to save both time and space in the computation. By carefully handling relation between each level and component of hierarchy, the proposed hierarchy based shape distribution features(HSD) should at least have same description power as the original shape distribution feature.

The hierarchical structure is built through an octree. Given a node $n_i$ of the octree, the basic idea is that I use combination of $n_i$'s children nodes to compute HSD. Take D2 feature for example, now we are computing D2 between $n_i$'s children nodes $n_i1$ and $n_i2$ which in fact is the D2 distance between two point clouds. Since it is likely that either $n_i1$ or $n_i2$ still contain a large number of points which make the computation back to the problem we face in the original shape distribution. So, in stead of computing exact D2 between two point clouds, I compute the expectation and the variance, then I simulate the histogram voted by the original D2 by using the expectation and variance.  

\begin{figure}[h]
\centerline{
\begin{tabular}{ccc} 
  \resizebox{0.3\textwidth}{!}{\rotatebox{0}{
  \includegraphics{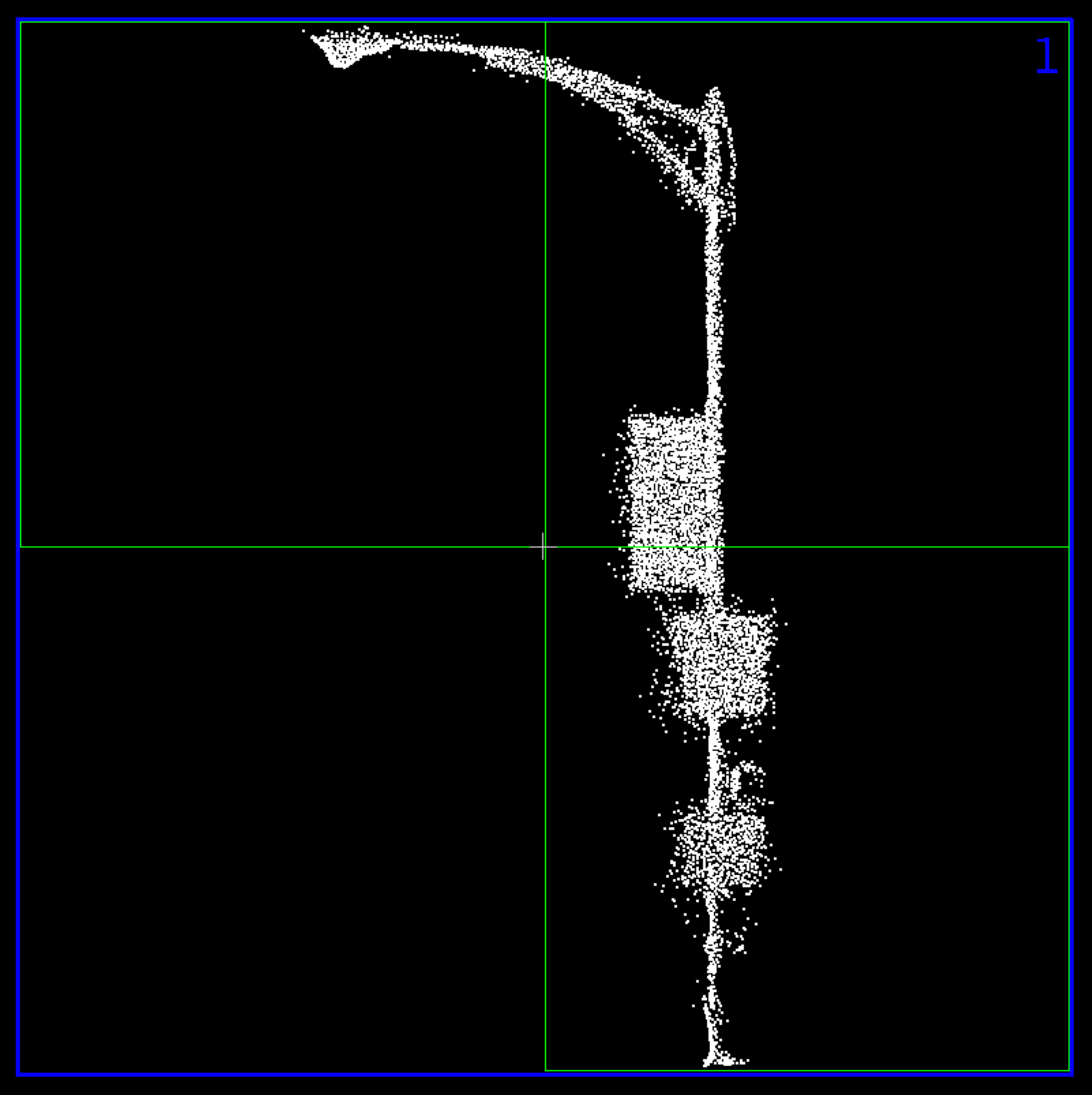}}}
  &
  \resizebox{0.3\textwidth}{!}{\rotatebox{0}{
  \includegraphics{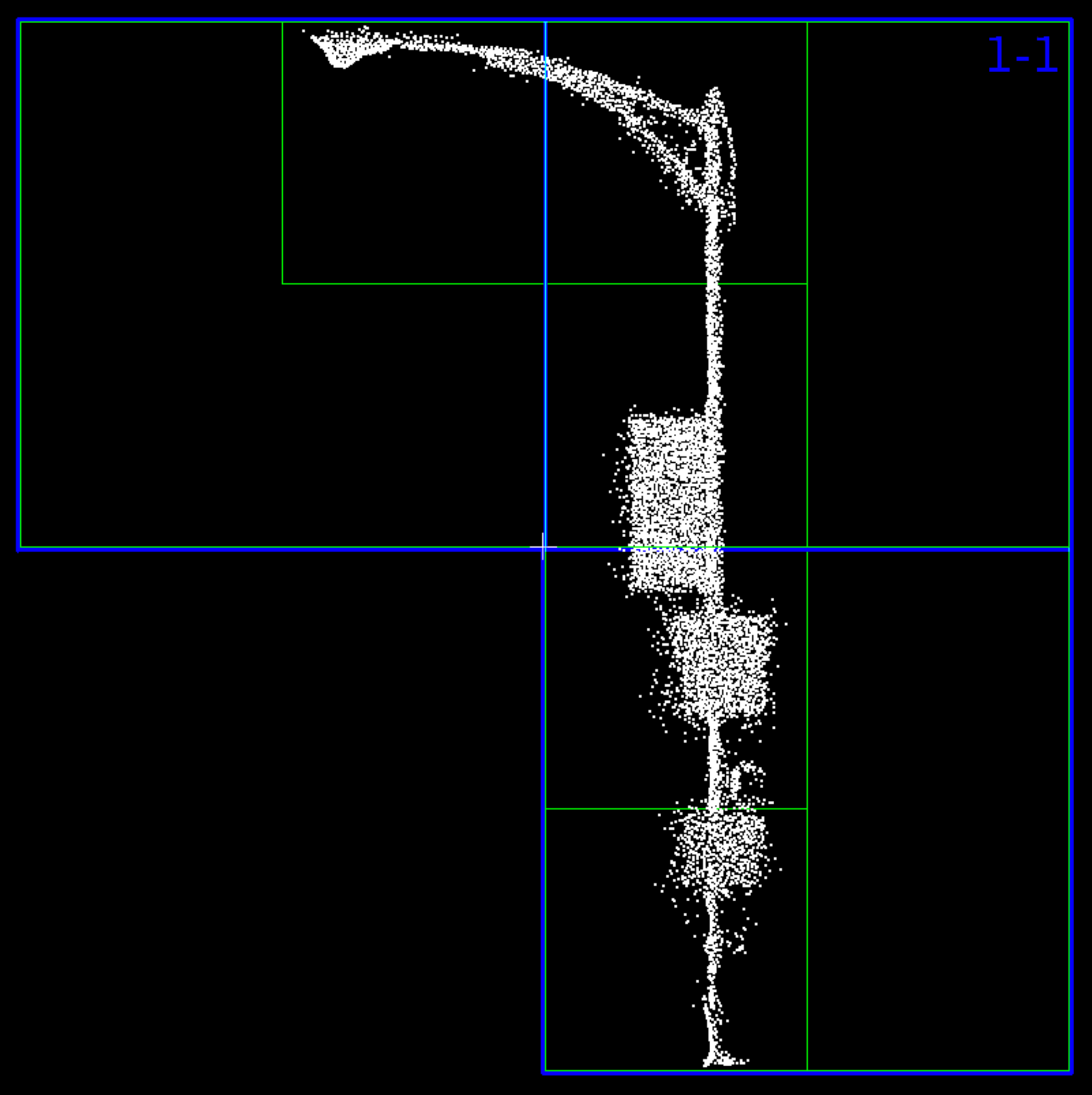}}}
  &
  \resizebox{0.3\textwidth}{!}{\rotatebox{0}{
  \includegraphics{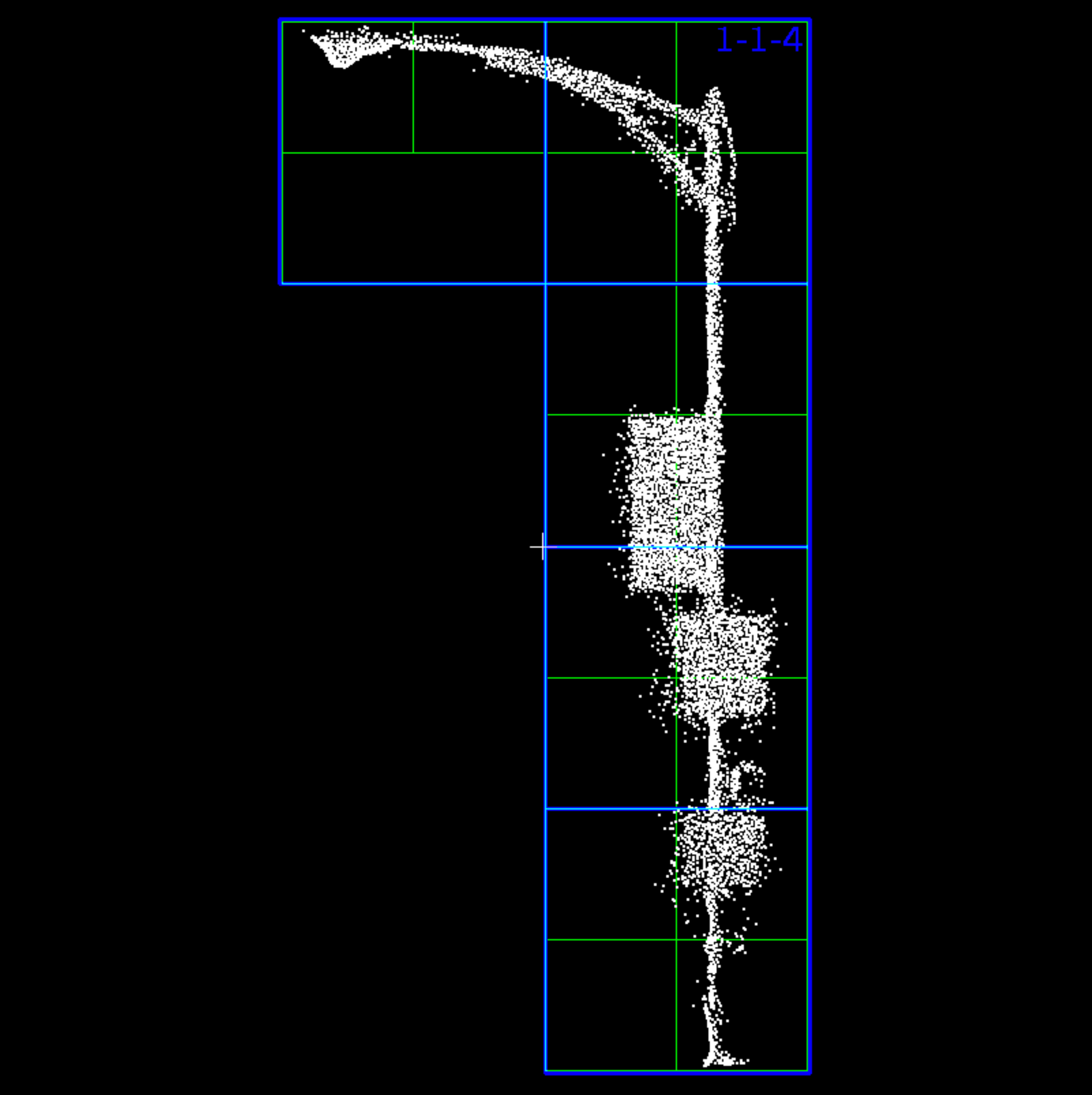}}}
  \\
  Level 1 & Level 2 & Level 3
\end{tabular}}
\caption{Computation of shape distribution feature using Octree structure.}
\label{fig: octree}
\end{figure}

\begin{figure}[h]
\centerline{
\begin{tabular}{c} 
  \resizebox{0.15\textwidth}{!}{\rotatebox{0}{
  \includegraphics{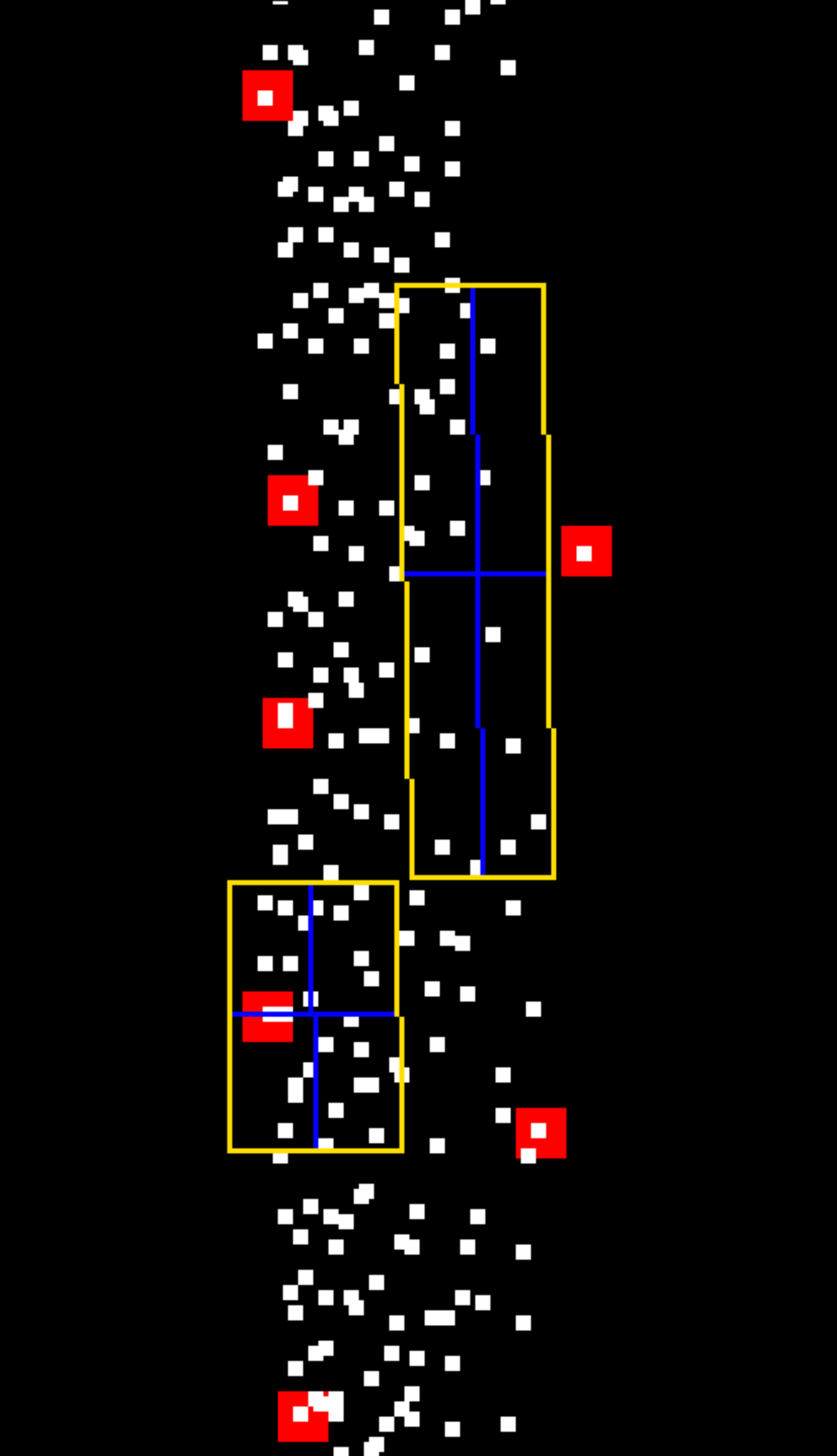}}}
  \\
\end{tabular}}
\caption{Using simplified point to compute shape distribution. The yellow box gives range of parameters in computation of integration.}
\label{fig: zoominmerged}
\end{figure}

However, using single expectation and variance to describe a point cloud is inaccurate and not enough. So, for a give point cloud within a node, I actually compute a down-sampled version of point cloud and compute expectations and variance of HSD for each point in the simplified point cloud.

The evaluation is done on a dataset I created from street view lidar. The dataset contains six categories of objects: Car (166), Garbage can (54), Pedestrian (220), Street light (82), Traffic light (26) and Tree (138).

I compute D2, A3, T3 and R3 features for all objects in the dataset, finally 4 histogram features are generated for each object. The evaluation then is done based on feature distance within the group and across the groups.

The measure the 4 histogram features differences between two object, we adapt and compare three strategies of computation of the difference: difference based on average distances, the biggest distances and the smallest distances among 4 histograms. 

\begin{figure}[h]
\centerline{
\begin{tabular}{ccc} 
  \resizebox{0.3\textwidth}{!}{\rotatebox{0}{
  \includegraphics{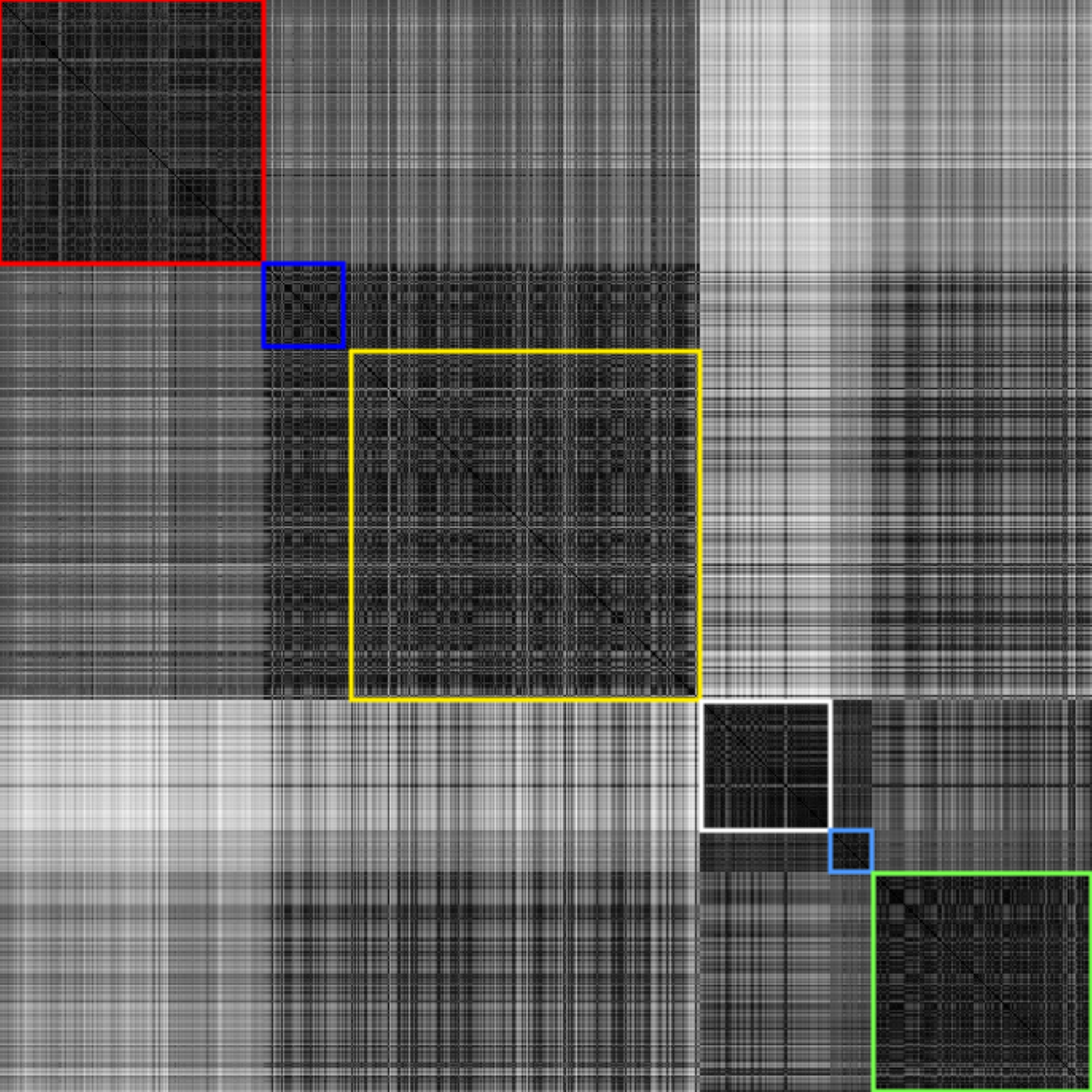}}}
  &
  \resizebox{0.3\textwidth}{!}{\rotatebox{0}{
  \includegraphics{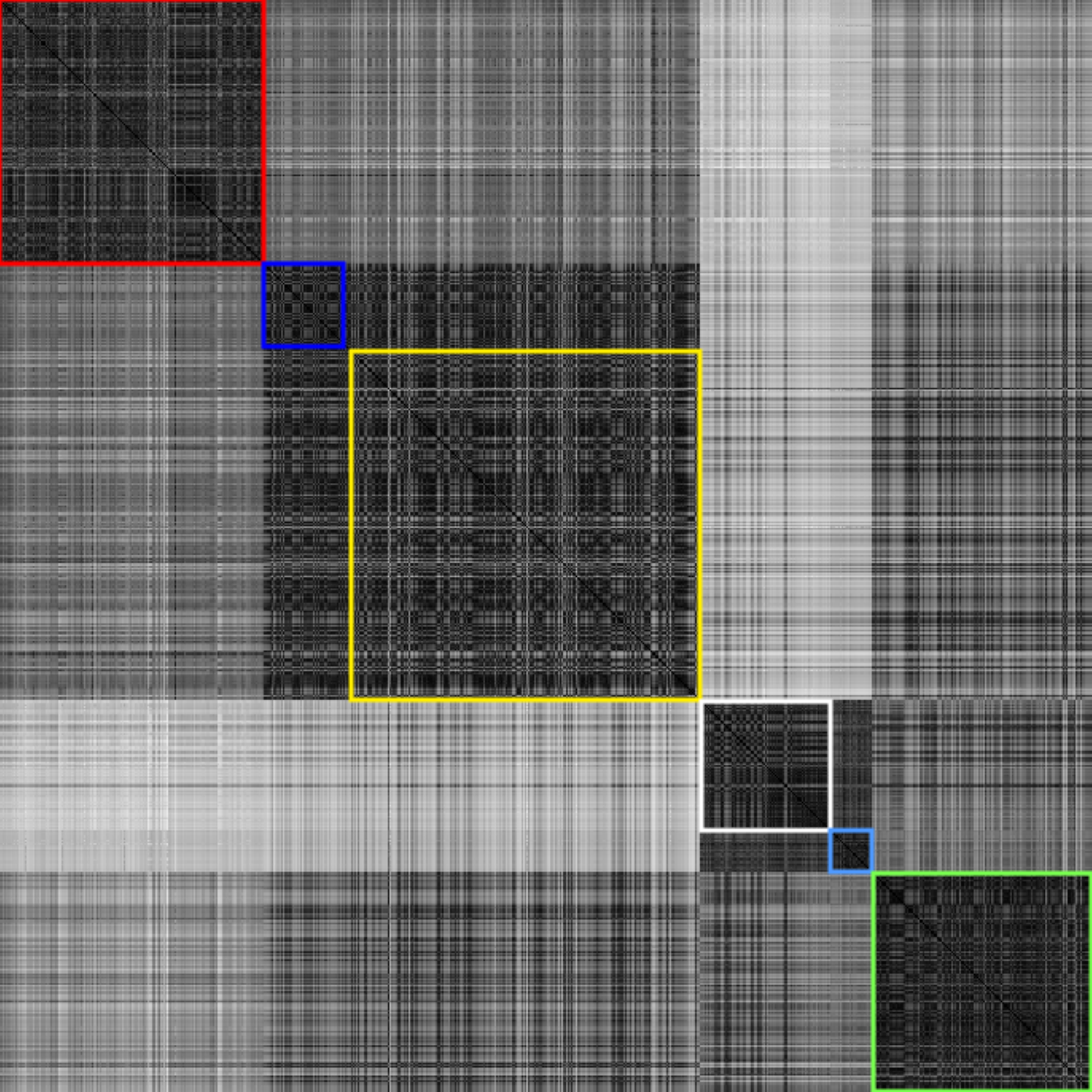}}}
  &
  \resizebox{0.3\textwidth}{!}{\rotatebox{0}{
  \includegraphics{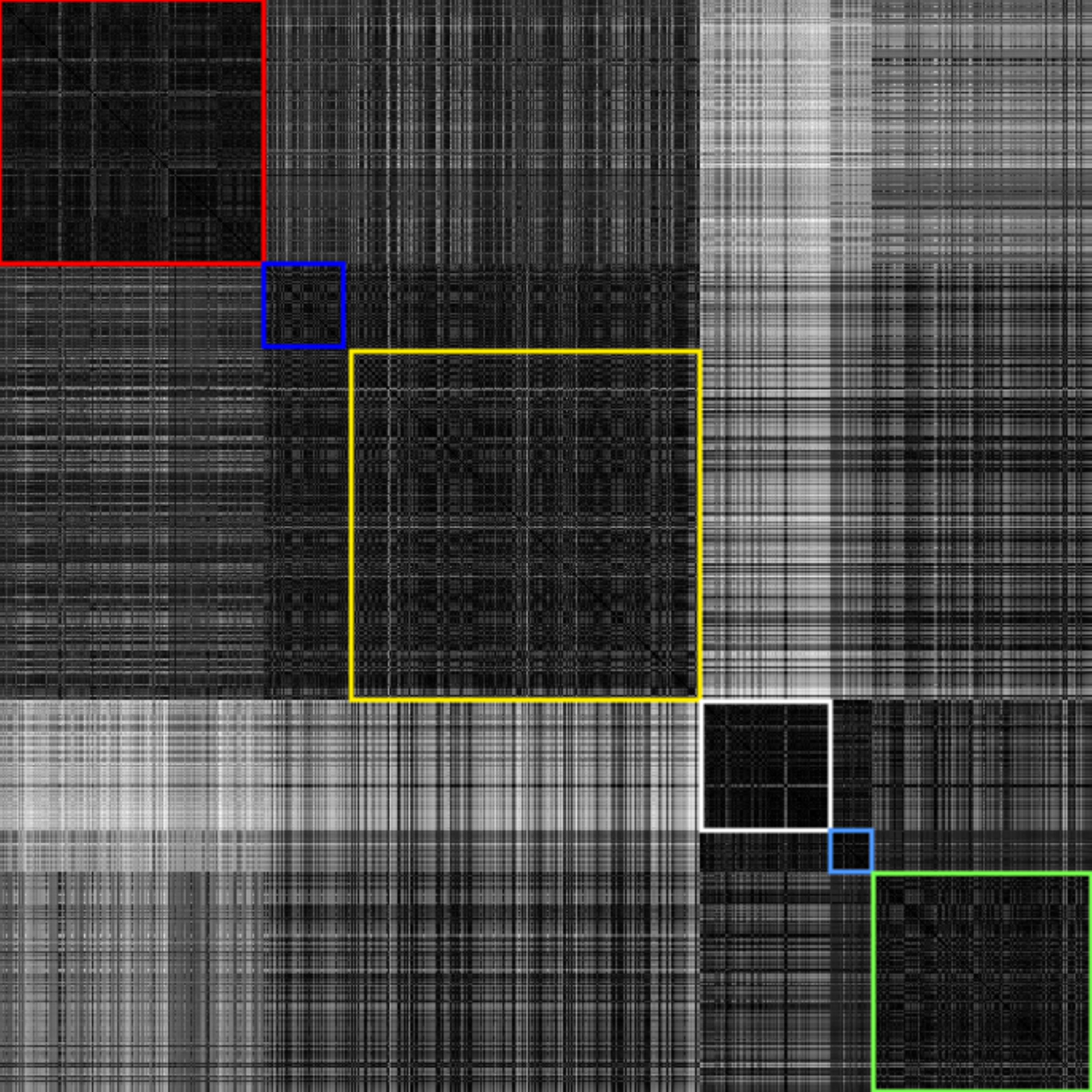}}}
  \\  
  Average dist. & Smallest dist. & Biggest dist.
  \\ 
  \resizebox{0.3\textwidth}{!}{\rotatebox{0}{
  \includegraphics{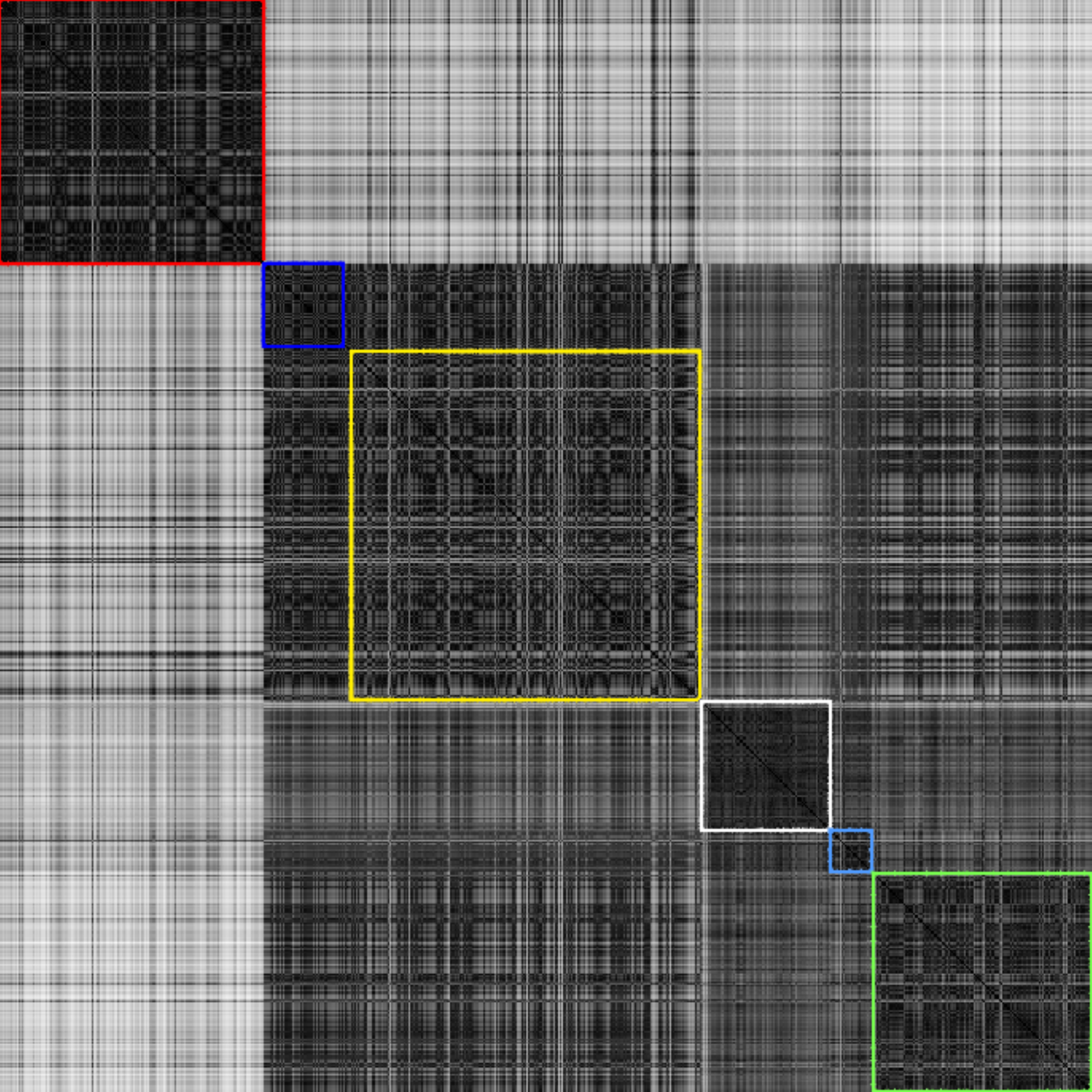}}}
  &
  \resizebox{0.3\textwidth}{!}{\rotatebox{0}{
  \includegraphics{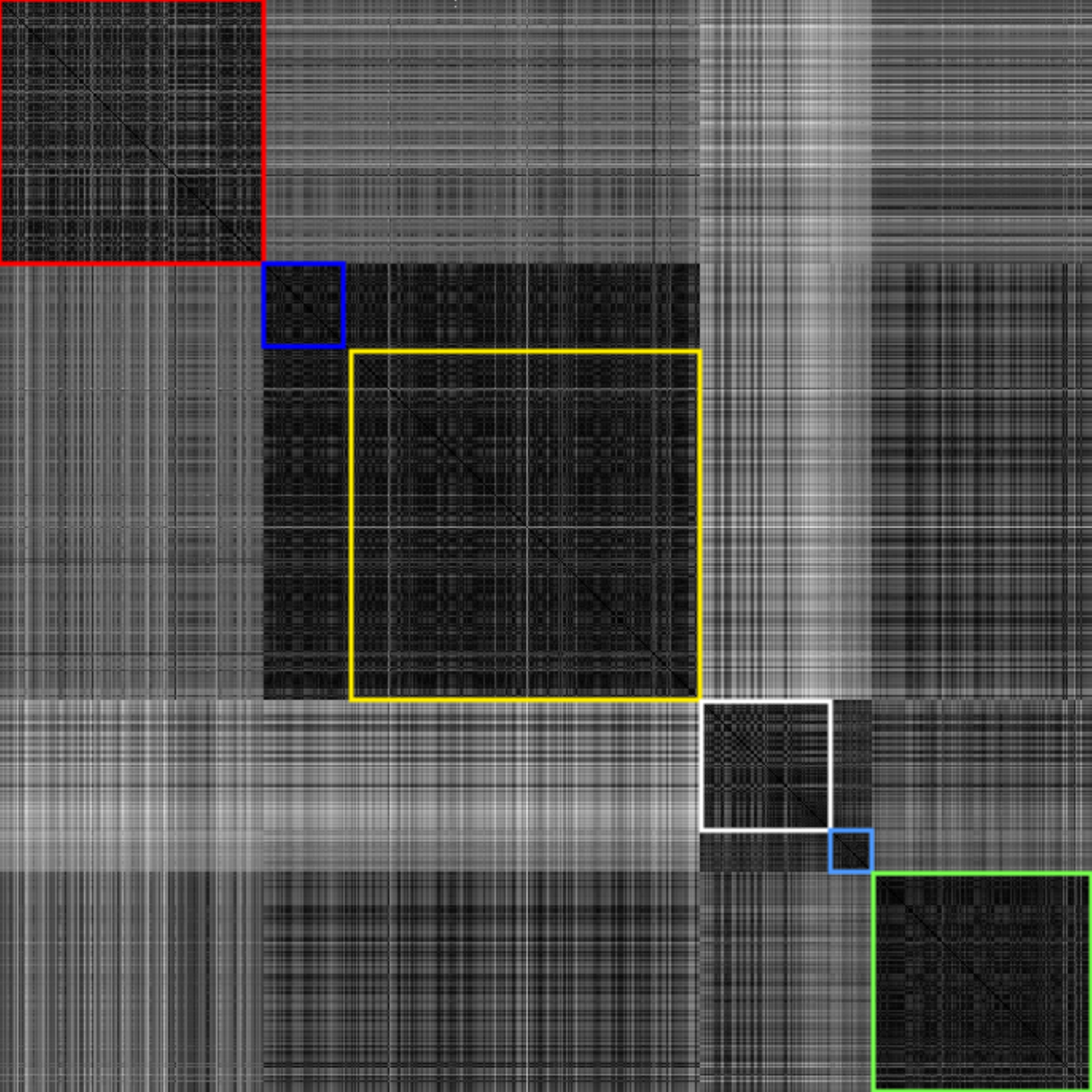}}}
  \\
  Original D2 feature. & HSD D2 feature.
  \\
\end{tabular}}
\caption{Distance matrix.}
\label{fig: distance matrix}
\end{figure}

The distance matrix are given in Figure \ref{fig: distance matrix}.

\begin{figure}[h]
\centerline{
\begin{tabular}{c} 
  \resizebox{0.8\textwidth}{!}{\rotatebox{0}{
  \includegraphics{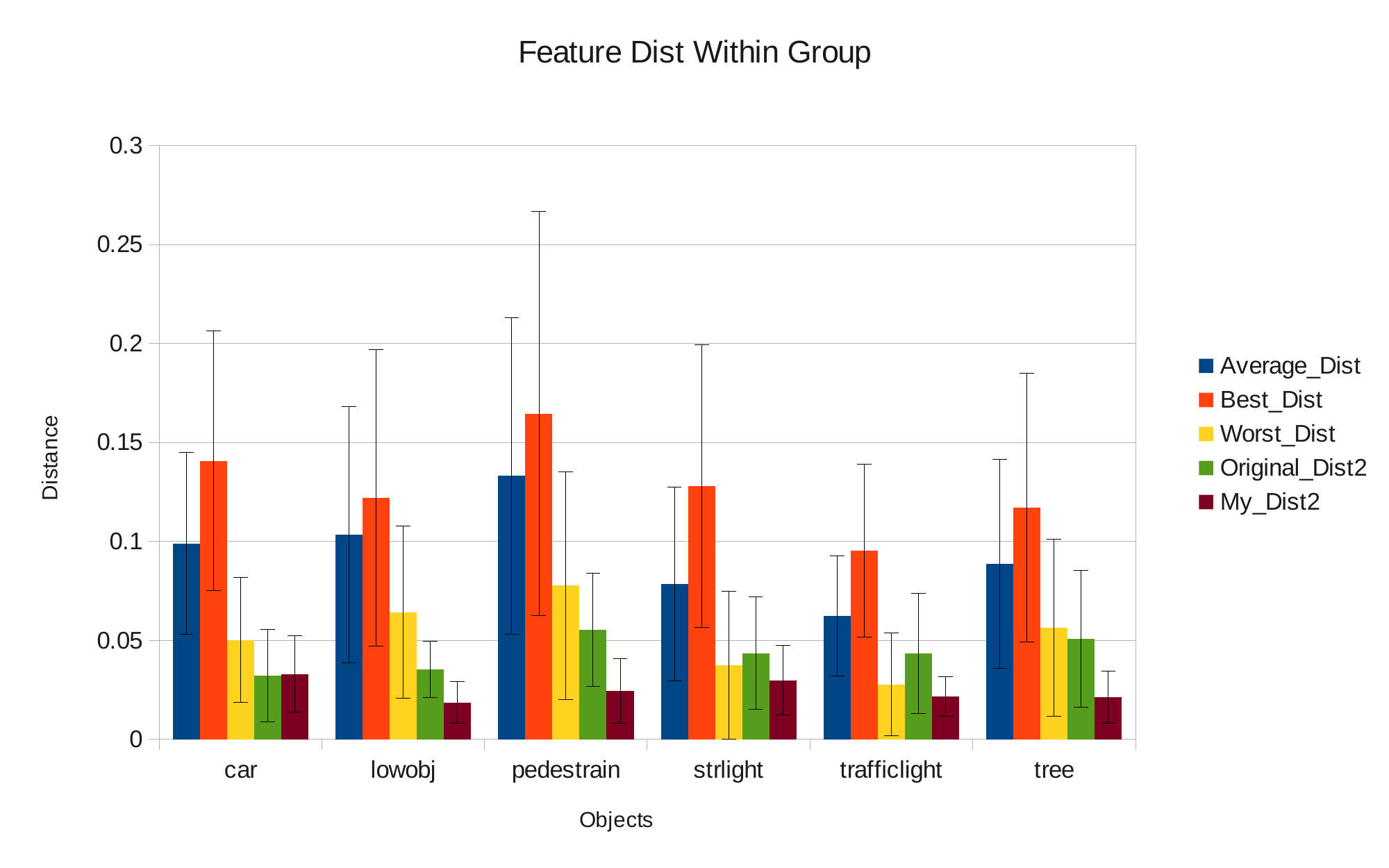}}}
  \\
  Distance in group
  \\
  \resizebox{0.8\textwidth}{!}{\rotatebox{0}{
  \includegraphics{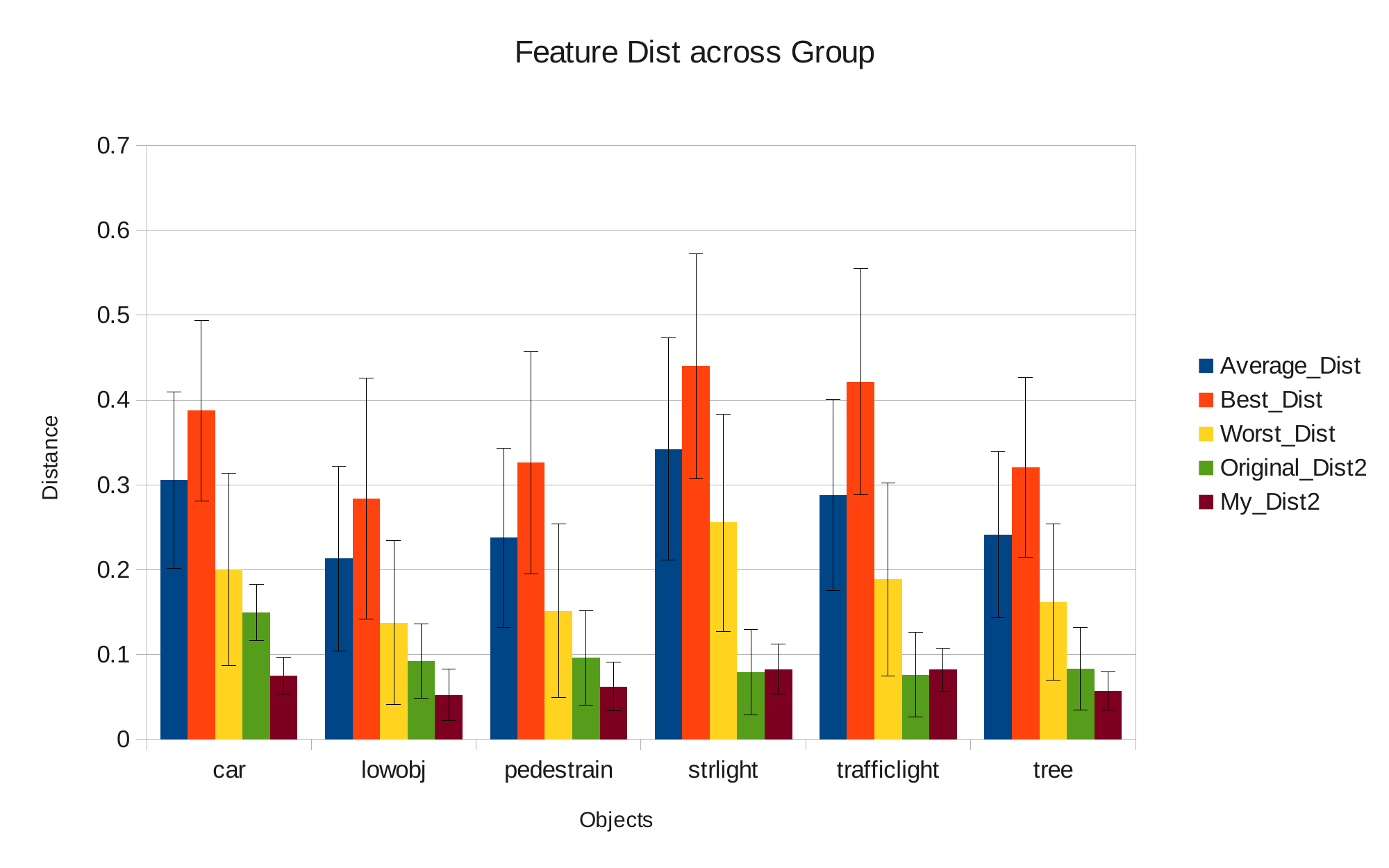}}}
  \\
  Distance across groups
  \\
\end{tabular}}
\caption{Distance difference within the object group.}
\label{fig: within the group}
\end{figure}

I also evaluate the mean and variance of feature difference within and across the group. The mean and variance within the group is show in Figure \ref{fig: within the group}, while the case of across the group is shown in Figure \ref{fig: across the group}.

\begin{figure}[ht]
\centerline{
\begin{tabular}{c} 
  \resizebox{0.8\textwidth}{!}{\rotatebox{0}{
  \includegraphics{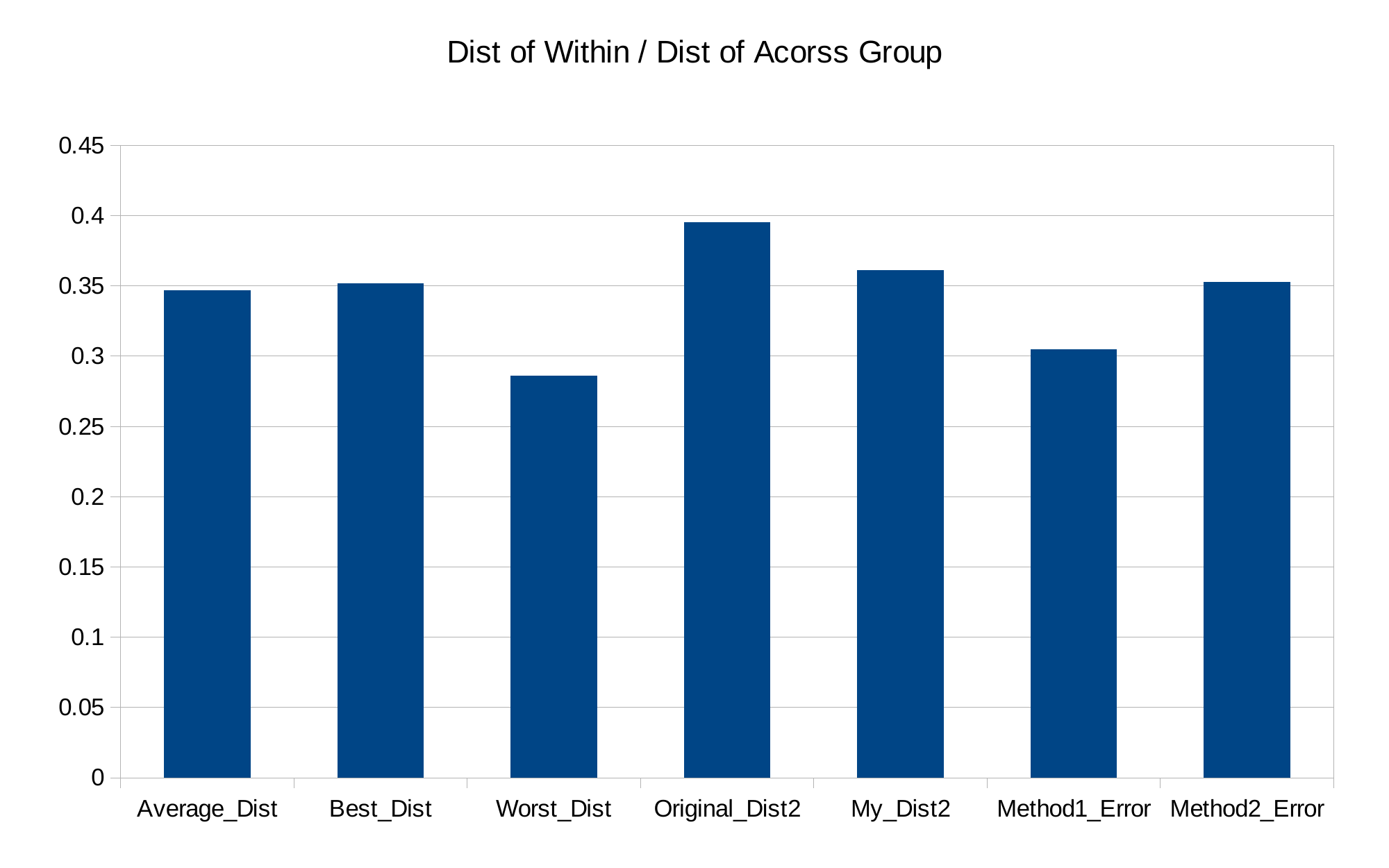}}}
  \\
\end{tabular}}
\caption{Ratio of distance within over across the groups}
\label{fig: ratio}
\end{figure}

Since we hope the distance among object feature within the group as small as possible, while the distance across other groups as big as possible, I compute the ratio between distance within and across the group. We hope this ratio to be as low as possible. The ratio of five metrics used are given in Figure \ref{fig: ratio}.

{\small
\bibliography{xibib}
\bibliographystyle{plain}
}

\end{document}